\documentclass[lettersize,journal]{IEEEtran}
\usepackage{amsmath,amsfonts}
\usepackage{algorithmic}
\usepackage{algorithm}
\usepackage{array}
\usepackage[caption=false,font=normalsize,labelfont=sf,textfont=sf]{subfig}
\usepackage{textcomp}
\usepackage{stfloats}
\usepackage{url}
\usepackage{verbatim}
\usepackage{graphicx}
\usepackage{cite}
\usepackage{booktabs}
\usepackage{mathrsfs}

\begin{document}

\title{Adapting Knowledge for \\ Few-shot Table-to-Text Generation}



\author{Zhixin~Guo,~\IEEEmembership{Student Member,~IEEE}, Mingxuan~Yan, Jiexing~Qi, Jianping~Zhou, \\ Ziwei~He, Guanjie~Zheng, Xinbing~Wang,~\IEEEmembership{Senior Member,~IEEE}, and Chenghu Zhou

\IEEEcompsocitemizethanks{\IEEEcompsocthanksitem Zhixin Guo, Mingxuan Yan, Jiexing Qi, Jianping Zhou, Ziwei He, and Xinbing Wang are with the School of Electronic Information and Electrical Engineering, Shanghai Jiao Tong University. \protect\\
E-mail: \{stjgzx, galaxy\_ymx, qi\_jiexing, jianpingzhou, ziwei.he, xwang8\}@sjtu.edu.cn

\IEEEcompsocthanksitem Guanjie Zheng is with John Hopcroft Center for Computer Science, Shanghai Jiao Tong University.  \protect\\
E-mail: gjzheng@sjtu.edu.cn

\IEEEcompsocthanksitem Chenghu Zhou is with the Institute of Geographic Sciences and Natural Resources Research, China.  \protect\\
E-mail: zhouchsjtu@gmail.com}
}





\maketitle
\begin{abstract}
Pretrained language models (PLMs) have made remarkable progress in table-to-text generation tasks. However, the lack of domain-specific knowledge makes it challenging to bridge the topological gap between tabular data and text, especially in real-world applications with limited resources. To mitigate the limitation of insufficient labeled data, we propose a novel framework: Adapt-Knowledge-to-Generate (AKG). The core insight of AKG is to adapt unlabeled domain-specific knowledge into the model, which brings at least three benefits: (1) it injects representation of normal table-related descriptions to bridge the topological gap between tabular data and texts; (2) it enables us to use large amounts of unlabeled domain-specific knowledge fully, which can alleviate the PLMs' inherent shortcomings of lacking domain knowledge; (3) it allows us to design various tasks to employ the domain-specific knowledge. Extensive experiments and analyses are conducted on three open-domain, few-shot natural language generation (NLG) data sets: Humans, Songs, and Books. Compared to previous state-of-the-art approaches, our model achieves superior performance in terms of both ﬂuency and accuracy as judged by human and automatic evaluations.
\end{abstract}


\section{Introduction}

\IEEEPARstart{G}{enerating} descriptive text from structured data \cite{gatt2018survey}, i.e., table-to-text generation, is an important research problem for various downstream natural language generation (NLG) applications. Some representative examples are question answering \cite{chen2021finqa},\cite{ghazvininejad2018knowledge},\cite{chen2020open}, dialog \cite{he2017learning}, report generation \cite{wiseman2017challenges},\cite{murakami2017learning},\cite{hasan2019clinical}, \cite{guo2023towards}, and biographical description \cite{lebret2016neural}, demonstrating the great potential of table-to-text generation for use in extensive real-world scenarios. \newline

The main challenge in table-to-text generation is the structural difference between the table and the natural language text. With the blossoming of deep neural networks, Pretrained Language Model (PLM)-based NLG systems have shown a remarkable ability to produce fluent text with informative content and have achieved state-of-the-art performance in many table-to-text tasks, such as WIKIBIO \cite{lebret2016neural}, RotoWire \cite{iso2019learning}, and ToTTo \cite{parikh2020totto}. However, these methods depend on a large training data set, and this data-hungry nature prevents neural models from being widely adopted for real-world applications. \newline

\begin{figure}[!t]
\centering
\includegraphics[scale=0.3]{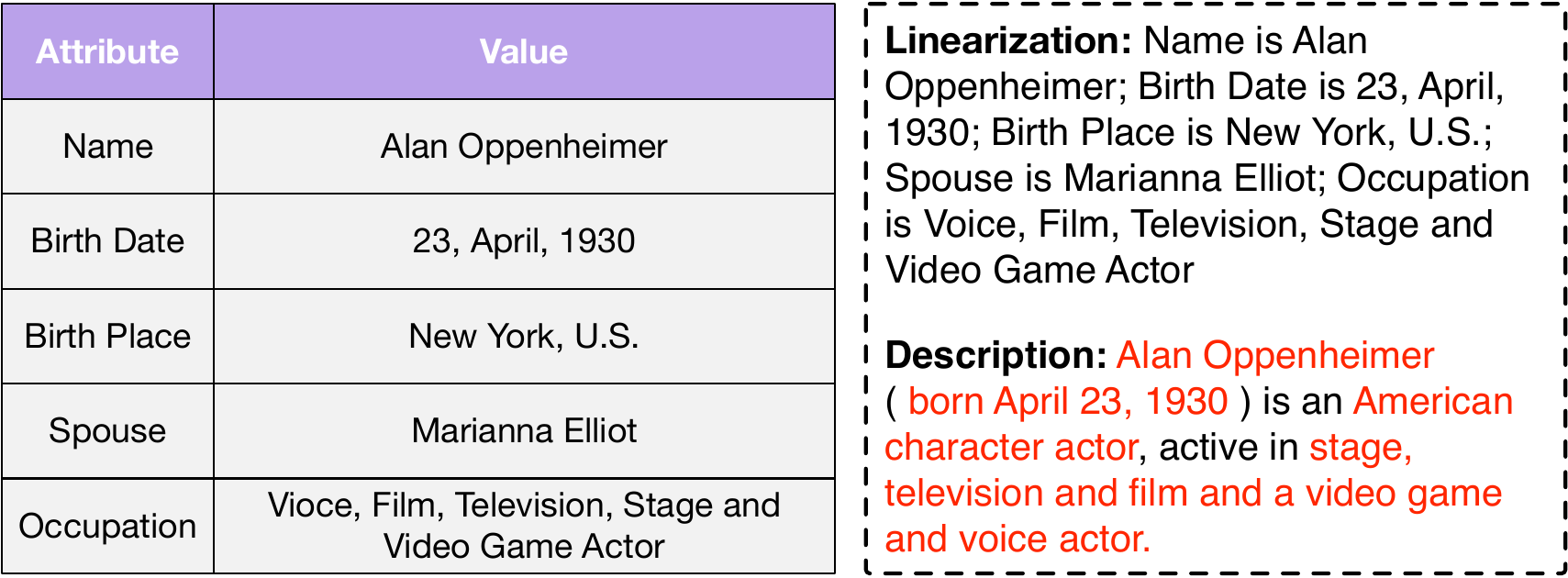}
\caption{An example of a table--text pair for few-shot table-to-text generation from the Humans data set. On the right-hand-side are a template of the key-value pair for table linearization and a description of the table. The red text indicates the content supported by the tabular data.}\protect
\label{TableSample}
\end{figure}

To mitigate this challenge, researchers are investigating various techniques to leverage the prior knowledge of PLMs fully, such as switch policy \cite{chen2020few} and table structure reconstruction \cite{gong2020tablegpt}. Recently, with the development of prompt learning, the ``prompt-tuning'' paradigm with PLMs has also been explored in table-to-text generation. Prefix-tuning \cite{li2021prefix} and prefix-controlled generation \cite{luo2022few} modify the encoder-decoder architecture with task-specific vectors as prompts, improving the efficiency of leveraging prior knowledge. These methods rely on fine-tuning the semantic knowledge and linguistic patterns learned from a large corpus during pretraining. \newline

Although fluency improves significantly, these methods always fabricate phrases unsupported by the table. Due to the limitation of insufficient training data, these methods are inadequate to capture the representation differences between the tabular data and descriptions. \cite{su2021few}, who first attempted the knowledge-augmentation method through a retrieval-based framework for providing related background information, significantly improved few-shot table-to-text generation. However, due to the input length limitation of their considered PLM, they only considered the top $n$-retrieved sentences, leaving out most of the information. In general, the main challenge, bridging the gap between the tabular data and text, is still underexplored under the few-shot condition due to the inherent shortcoming of PLMs lacking domain-specific knowledge. \newline
 
To address this problem, we propose the AKG framework, which breaks through the bottleneck of lacking domain-specific knowledge. Compared to the previous study \cite{su2021few}, we have improved the efficiency of domain knowledge usage without adding additional resources. We leverage an unlabeled corpus pertinent to table contexts to craft and refine table-related prompt templates. These templates are then seamlessly integrated into the AKG framework employing a modularized pretraining approach, enabling the dynamic adaptation of domain-specific insights. To evaluate our approach comprehensively, we evaluate the proposed method on a multidomain table-to-text data set. Compared with previous state-of-the-art approaches, our method achieves remarkable improvement in fluency and faithfulness of the generated contents as judged by human and automatic evaluations. Moreover, we also perform extensive ablation studies of AKG. In particular, the results also illustrate that our model outputs are highly faithful and fluent. In short, our contributions can be summarized as follows: 
\begin{itemize}
\item We propose a novel framework AKG for few-shot table-to-text generation that attempts to alleviate the insufficient-data limitation to bridge the topological gap between tabular data and text. AKG enables the model to make full use of a large domain-specific knowledge corpus.
\item We design an effective modularized pretraining strategy for adapting the prompt templates to inject table-related representation and domain-specific knowledge. The modularized pretraining strategy effectively integrates various tasks to employ the domain-knowledge.
\item We conduct extensive experiments on a multi-domain table-to-text data set encompassing three distinct areas: Humans, Books, and Songs. Both automatic and human evaluations report state-of-the-art performance.
\end{itemize}

\section{Related Work}
\begin{figure*}[!t]
\centering
\includegraphics[scale=0.35]{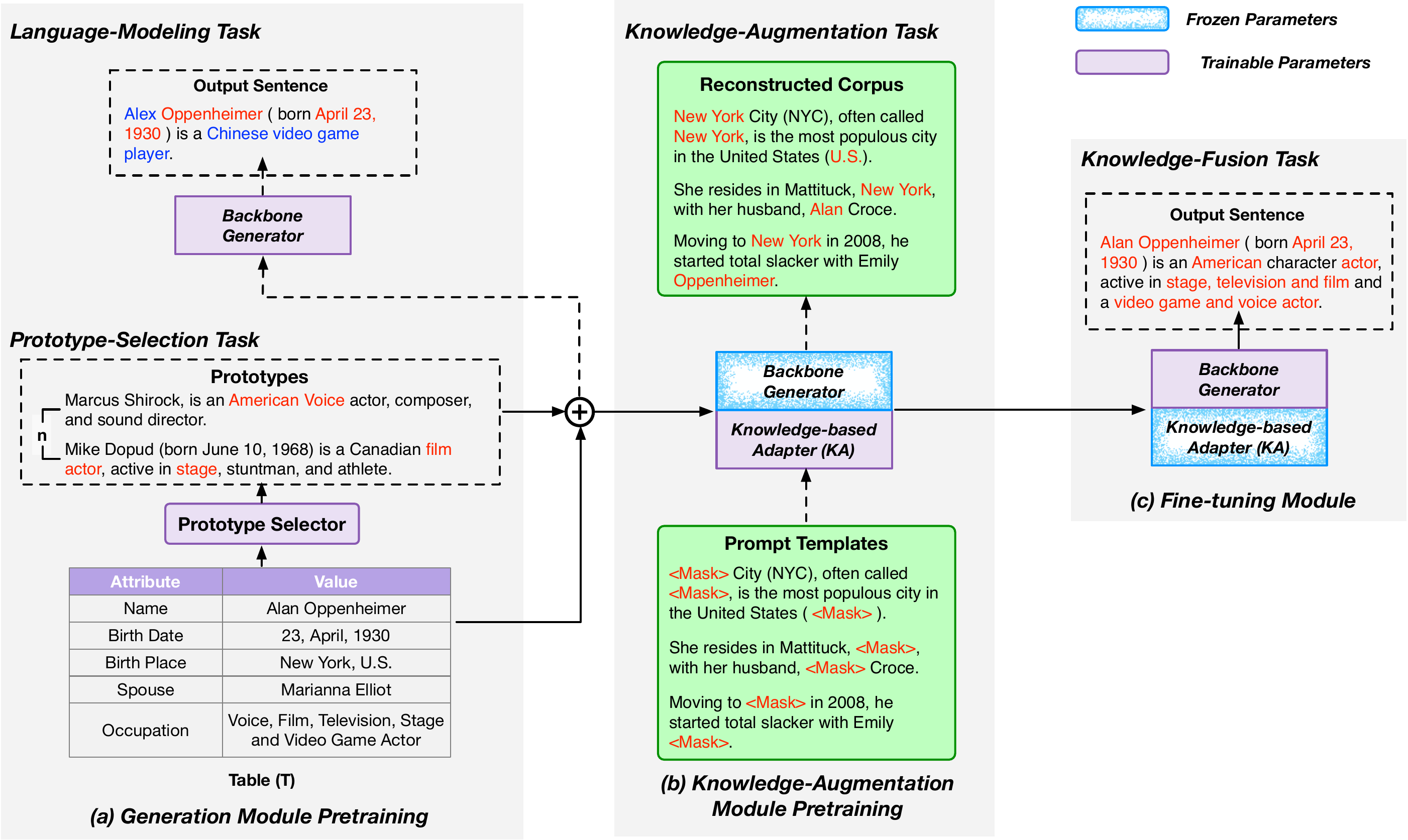}
\caption{An overview of the AKG framework. We propose a modularized pretraining strategy that targets AKG. The pretraining strategy consists of three modules: (a) Generation Module, (b) Knowledge-Augmentation Module, and (c) Fine-tuning Module. Throughout the generation module pretraining process, we divide the generation module into two tasks: a prototype-selection task and a language-modeling task. The prototype-selection task selects related prototypes to guide the generation of PLMs, and the language-modeling task employs a PLM as the backbone generator to generate fluent sentences. We pretrain these two tasks separately. We insert the Knowledge-based Adapter (KA) after the final layer of each encoder and decoder and adapt the generated prompt templates through the KA during the pretraining of the knowledge-augmentation module. All the parameters except those of the KA are frozen when pretraining the knowledge-augmentation module. The knowledge-augmentation module brings at least two benefits: (1) it enables us to use the large amounts of unlabeled domain-specific knowledge fully, which can alleviate the PLM’s inherent shortcomings of lacking domain knowledge; and (2) it allows us to design various tasks to employ the domain-specific knowledge. Finally, we fine-tune the pretrained modularized model on three data sets: Humans, Books, and Songs. Throughout the fine-tuning process, the parameters of the KA are frozen to retain the learned knowledge pretrained from the knowledge-augmentation module. The red text indicates the tabular data supporting the fact, and the blue text indicates that the (incorrect) fact conflicts with the information in the original table.}\protect
\label{Framework}
\end{figure*}

As it is an essential objective in many real-world scenarios, researchers have investigated NLG from tabular data for many years. Early conventional generation systems followed the pipeline paradigm, such as macro/micro planning \cite{reiter1997building} and template-based content selection \cite{liang2009learning},\cite{walker2001spot},\cite{lu-etal-2009-natural}. Such pipeline approaches significantly rely on feature engineering and template design. Later works, with the blooming of deep neural networks, employed neural-based methods and achieved remarkable performance in table-to-text generative challenges like WIKIBIO \cite{liu2018table}, RotoWire \cite{iso2019learning}, WebNLG \cite{gardent2017webnlg}, E2E \cite{novikova2017e2e}, and ToTTo \cite{parikh2020totto}. Researchers explored optimizing deep neural networks to bridge the gap between structured data and text, such as copy mechanism \cite{see2017get} and content-aware generation \cite{gong2020tablegpt}. However, such methods rely on a tremendous amount of labeled data. Towards targeting real-world applications, studies of low-resource generation from structured data have gained increased attention. Zero-shot learning for question generation from knowledge graphs \cite{elsahar2018zero} and open-domain few-shot table-to-text generation \cite{chen2020few} illustrate that PLMs suffer from limited labeled data in a few-shot setting due to the lack of domain-specific knowledge. \newline

To alleviate labeled data dependency, researchers attempted to modify the architecture of the PLM to improve the efficiency of using prior information. The Semantically Conditioned Variational AutoEncoder (SCVAE) \cite{tseng2018variational} was proposed for semantic-controlled generation. Learning domain-invariant representations achieved impressive performance, allowing for improvements of the low-resource Semantically Conditioned Long Short-Term Memory (SCLSTM) settings. Similarly, \cite{tran2018dual} employed the same idea of designing a refinement adjustment LSTM-based component to select and control the semantic information better. Apart from these, algorithms represented by Model-Agnostic Meta-Learning (MAML) \cite{finn2017model} for improving parameter efficiency have also been investigated in NLG. For example, \cite{mi2019meta} proposed Meta-NLG for task-oriented dialogue systems based on the MAML algorithm by defining meta-tasks for adapting to low-resource NLG tasks. \cite{yao2021knowledge} suggested structured meta-learning for knowledge-aware algorithms for text classification. However, these explorations can barely be generalized directly to table-to-text generation due to different application scenarios.  \newline

Recently, prompt tuning has achieved impressive success in table-to-text NLG. Preﬁx-tuning \cite{li2021prefix} prepends learned task-specific vectors and performs well. Prefix-controlled \cite{luo2022few} generators use prefix vectors as a planned signal to guide and control the output of PLMs. Plan-then-generate \cite{su2021plan} uses a content planner to generate key facts as a prompt signal to instruct generation. Although these approaches reduce the number of parameters in the model compared to previous methods following the prompt-tuning paradigm, the lack of domain-specific knowledge still needs to be addressed. \cite{su2021few} demonstrated a retrieval-based generation framework to provide background knowledge and commonsense reasoning information. However, due to the input length limitation of the considered PLMs, their method only takes the top $n$-retrieved sentences, leaving out most of the information. Unlike previous works, we propose AKG for table-to-text generation and focus on improving fluency and accuracy with a modularized pretraining strategy for injecting table-related representation and domain-specific knowledge.
\section{Preliminaries}

In this section, we brieﬂy introduce the relevant research lines of our work: table-to-text generation, the Prototype-To-Generate (P2G) framework, and the parameter-efficient adapter.\newline

\textbf{Table-to-text generation} 
The objective of the table-to-text task is to synthesize coherent and accurate descriptions in natural language that accurately reflect the contents of a given table. In the context of this research, the training data set is represented as $D=\left\{(T, R)_i\right\}_{i=1}^{|D|}$, where each tuple $(T, R)_{i}$ comprises a linearized table $T$ and its associated reference text $R$. The table $T$ is composed of one or more tabular data elements, denoted as $T= \left\{t_{1},\cdots, t_{|n|}\right\}$. Each element $t_{i}=\left\{a_{i}; v_{i}\right\}$ within the table represents an attribute-value pair, with $a_{i}$ and $v_{i}$ potentially being a string/number, a phrase, or a complete sentence. Subsequently, $R=\left\{r_1,\cdots,r_{|m|}\right\}$ signifies the set of reference texts, with each $r_i$ providing a narrative description that corresponds to the table $T$. This framework facilitates the generation of natural language descriptions that are both fluent and faithful to the tabular data presented. \newline

Unlike the rich corpus data used for PLMs, tabular data contain complex topology structures with few narrative descriptions, which are far from natural language representations. Fig. \ref{TableSample} illustrates an example of tabular data from the Humans data set. According to the experimental results of \cite{gong2020tablegpt}, the tabular data format significantly enhances generation performance. In this paper, to make the tabular data more compatible with the internal representations of PLMs, we leverage a template for table linearization. As shown in Fig.~\ref{TableSample}, we use the key-value pair template ``name: Alan Oppenheimer" as ``Name is Alan Oppenheimer", then stack all key-value pairs to form a sentence, that is, ``Name is Alan Oppenheimer; Birth Date is 23, April, 1930; Birth Place is New York, U.S.; ...".

\textbf{P2G framework}
Previous work by P2G \cite{su2021few} achieved impressive results. \cite{su2021few} designed a retrieval-based framework by retrieving domain-related information from an external information retrieval (IR) system as prototype memory to guide the PLM in generating fluent and faithful sentences. Given input tabular data $T$ with its corresponding reference text $R$, P2G first retrieves the $n$ most related corresponding sentences from the IR system and generates prototype memory according to the semantic features of the candidates. Then, the method concatenates the prototype memory with the tabular data as input to the PLM to produce corresponding descriptions. As the method of \cite{su2021few} yielded reliable achievements in improving fluency and faithfulness, we employ a prototype-selection model to generate a plan to guide generation. \newline

\textbf{Parameter-efficient adapter}
The investigation into the use of adapters, as delineated in the studies by \cite{houlsby2019parameter}, \cite{hulora}, \cite{liufew}, and \cite{hu2021lora}, has attracted considerable interest in the field of transfer learning. This interest is primarily due to the demonstrated effectiveness of adapters in optimizing parameter utilization within diverse applications. Notably, their utility extends to enhancing the performance of task-oriented dialogue systems, as evidenced by the contributions of \cite{qin2023modularized} and \cite{emelin2022injecting}. Distinct from the conventional application of adapters across multiple tasks, our approach necessitates an efficient, architecture-neutral plug-in model capable of seamlessly integrating prompt and domain-specific knowledge to facilitate the generation process. To fulfill these criteria, we have adopted the method proposed by \cite{houlsby2019parameter} as the foundation for our knowledge-based adapter (KA), incorporating it within our generative framework as a knowledge-augmentation module. 

\section{Methodology}

This study introduces a novel approach to few-shot table-to-text generation by presenting a knowledge-augmentation methodology that leverages the AKG framework to refine prompts. Fig.~\ref{Framework} illustrates the comprehensive architecture of our proposed method. Within this framework, we integrate a prototype-selector, as described in \cite{su2021few}, which guides the PLM during the generation phase. Moreover, we develop a knowledge-augmentation task designed to enrich the model with prompt templates and domain-specific knowledge via the Knowledge Adapter (KA). This KA is strategically positioned after the terminal layer of each encoder and decoder in the backbone generator, as depicted in Fig.~\ref{Framework}. Our approach is characterized by a modularized pretraining strategy that segments the table-to-text generation process into three distinct modules: generation, knowledge augmentation, and fine-tuning. This modular structure facilitates the model's comprehensive exploitation of a vast domain-specific corpus and allows each submodule to engage in specialized pretraining tasks. These tasks are tailored to bolster the generative capabilities of the model, thereby enhancing its performance and adaptability in generating text from tabular data. \newline

As illustrated in Fig.~\ref{Framework}(a), our methodology employs a modular approach to the generation module, integrating both prototype-selection and language-modeling tasks. The objective of the prototype-selection task is to discern and extract relevant prototypes from an unlabeled corpus that align with the contextual requirements of the table. Simultaneously, the language-modeling task utilizes a PLM to generate cohesive sentences from linearized table formats. This dual-task strategy addresses and mitigates the structural discontinuity between tabular data and narrative descriptions, facilitating a seamless transition from tabular data representation to natural language interpretation. \newline

To enhance the accuracy of the generated content, our methodology is refined by pretraining a knowledge-augmentation module, which integrates table-specific prompt templates via the KA. This module undertakes a pivotal knowledge-augmentation task aimed at redefining prompt templates, which is achieved by reconstructing the prompt templates, effectively overcoming the inherent deficiencies of PLMs through the strategic employment of a vast corpus of unlabeled, domain-specific knowledge.  As shown in Fig.~\ref{Framework}(b), the knowledge-augmentation module first freezes the backbone generator and further pretrains the KA independently. Specifically, pretraining of the knowledge-augmentation module relies on reconstructing the prompt templates extracted from the unlabeled domain-specific corpus, distinct from the generation task. Such a modularization strategy results in at least three advantages: (1) each module can be easily integrated; (2) it allows the integration of different types of generation-support tasks; and (3) it enables the model to make full use of any unlabeled domain-specific corpus. \newline

As shown in Fig.~\ref{Framework}(c), we introduce a fine-tuning module for fusing the linguistic and semantic patterns and the augmented knowledge by pretraining the generation and knowledge-augmentation modules separately. Throughout the fine-tuning process, we freeze the parameters of the KA to retain the augmented knowledge. \newline

\subsection{Generation Module Pretraining}
\subsubsection{Prototype-Selection Task}
We employ a prototype retriever to select prototypes that relate to the input tabular data from the IR system. The task of prototype selection is to predict the similarity between the tabular data and the retrieved prototypes. Given the input tabular data, $T$, with the reference text, $R$, the retriever retrieves the $n$ most related corresponding prototype, $P$, from the IR system, $B$. Each candidate sentence of $B$ is defined as $b$ and each retrieved candidate set is defined as $B^{\prime}$. We utilize a Bidirectional Encoder Representations from Transformers (BERT)-based model \cite{devlin2018bert} to get the representation of the prototype and evaluate its similarity to the target table, $T$. The similarity score is denoted as $f(T, b)$, and $P$ is then defined as:
\begin{equation}
    P=\underset{B^{\prime} \in B,\left|B^{\prime}\right|=n}{\arg \max } \sum_{b \in B^{\prime}} f(T, b) .
\label{prototype-selection}
\end{equation}

It is computed by the linear projection of the average embedding of concatenated text, [T:b], by BERT. In order to select the most related candidate sentences, we utilize the hinge loss-based objective during the training process. Given target table $T$ with reference text $R$ and the retrieved candidate set $B^{\prime}$, the learning objective is defined as:
\begin{equation}
    L_{PS} = \sum_{j=1}^{k}\max{(0, 1-f(T, R) + f(T, b_j))} ,
\label{prototype-object}
\end{equation}
where $b_j \in B^{\prime}$ and $k$ is the number of text candidates sampled from $B^{\prime}$.

\subsubsection{Language-Modeling Task}
The language-modeling task aims to train the PLM to generate sentences that describe the tabular data. The AKG framework is model agnostic. Thus the backbone generator can be any generation model. Our experiments utilized BART-large \cite{lewis2020bart}, an encoder--decoder architecture transformer, as our backbone generator for its remarkable performance in generative challenges. Given the structured data, $T$, the prototype, $P$, and the reference text, $R$, the learning objective of the sequence generator is the cross-entropy loss, deﬁned as Equation~\ref{lmt}:
\begin{equation}
    L_{LM} = - \sum_{i=1}^{|R|} \log{P_{\mathscr{D+A}}}{(R_{i}|R_{<i}; \mathscr{E+A}([P:T]))} ,
\label{lmt}
\end{equation}

where $\mathscr{E+A}$ and $\mathscr{D+A}$ denote the configurations where the KA is integrated subsequent to the encoder and decoder, respectively. While the proposed method is agnostic to the choice of the particular PLM, we leave such validation for future work.

\subsection{Knowledge-Augmentation Module Pretraining}
\begin{figure}[!t]
\centering
\includegraphics[scale=0.33]{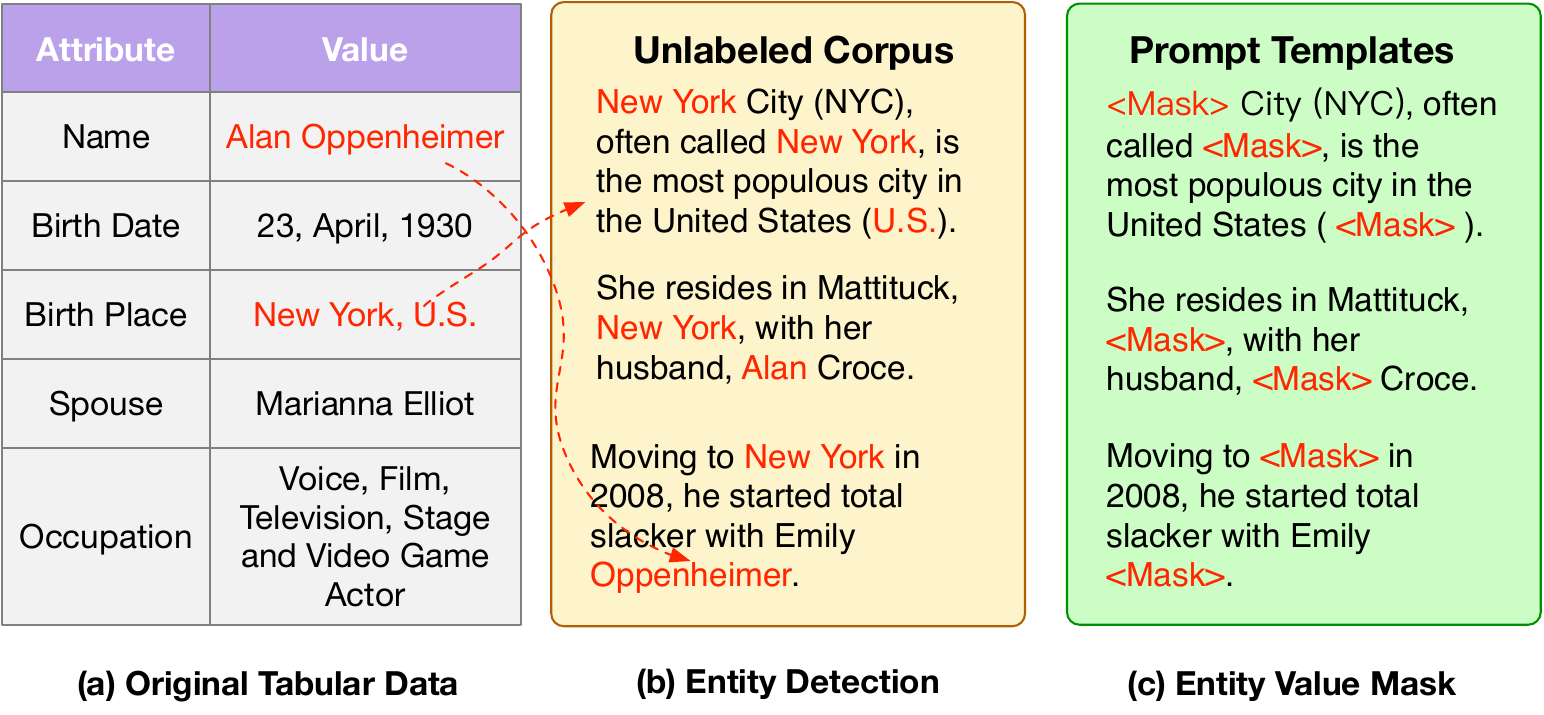}
\caption{Illustration of prompt generation. (a) The original tabular data, (b) illustration of the entity-detection process, and (c) the entity value mask process.}
\label{promptgen}
\end{figure}

\subsubsection{Prompt Generation}
The main idea of generating prompts is to replace the entities associated with the tabular data in order to reconstruct the knowledge representation. Unlike data augmentation in other areas, we require the augmented knowledge to follow two crucial rules: (1) it must contains table-related factors to bridge the structural gap between the tabular data and the texts the data represent; and (2) the domain-specific knowledge is injected to solve the insufficient training data problem. As shown in Fig.~\ref{promptgen}, the prompt-generation process consists of two steps, \emph{entity detection} and \emph{entity value mask}:

\begin{itemize}
    \item \textbf{Entity detection}. We first detect all entities and their attributes related to the tabular data in the unlabeled corpus provided by \cite{su2021few}. This process is accomplished by systematically matching each word's string in the table against individual words within the corpus. For example, ``New York, U.S." and ``Alan Oppenheim" are detected, which are shown in Fig.~\ref{promptgen}(b).
    \item \textbf{Entity value mask}.  We apply an entity value mask to generate prompts. As shown in Fig.~\ref{promptgen}(c), we replace ``New York" with ``\textless Mask\textgreater".
\end{itemize}

\subsubsection{Knowledge-Augmentation Task}
We strategically incorporate the KA subsequent to the final layer within the backbone generator's encoder and decoder. This precise integration facilitates the injection of prompts, enhancing the model's ability to assimilate and process the relevant knowledge. The intuition of such a design is that the pluggable knowledge adapter satisfies the lightweight, model-agnostic requirement and is adequate for fine-tuning entirely new tasks. The knowledge adapter consists of a down-projection matrix, $W_{down}$, and an up-projection matrix, $W_{up}$, as shown in Equation~\ref{kat}:

\begin{equation}
h \leftarrow W_{up} \cdot (W_{down} \cdot h)+r ,
\label{kat}
\end{equation}

$W_{down}$ projects the hidden states, $h$, into the lower-dimension $d_{bottleneck}$ and $W_{up}$ projects back into the original dimension of hidden states with a residual connection, $r$. \newline

The initialization of the backbone generator's parameters is conducted using the parameters derived from the antecedent generation module pretraining phase. All parameters except the KA are frozen during the pretraining of the knowledge-augmentation module, which allows the KA to retain the knowledge learned from the knowledge-augmentation task. Given the masked prompts $B = \left\{b_{1}, b_{2}, \cdots, b_{i} \right\}$ and the target sentence $\bar{B}=\left\{\bar{b}_{1}, \bar{b}_{2}, \cdots, \bar{b}_{i}\right\}$, the distribution is $P_{K}(\bar{B}|B)$ and the learning objective is defined as Equation~\ref{equation2}:
\begin{equation}
    L_{K} = - \sum_{i=1}^{|\bar{B}|} \log{P_{K}}({\bar{B}_{i}|\bar{B}_{<i}; B}) .
\label{equation2}
\end{equation}

\subsection{Knowledge Fusion through the Fine-Tuning Module}
Finally, following the distinct pretraining of the language-generation and knowledge-augmentation modules, we delineate fine-tuning the pretrained system via the knowledge-fusion task. This task is designed to refine the integrated model by leveraging the insights from both pretraining modules. The initialization of the parameters for both the backbone generator and the KA is executed by employing the parameters obtained from the preceding phase of knowledge-augmentation module pretraining. Throughout the knowledge-fusion phase, the parameters of the KA are meticulously frozen to safeguard the integrity of the enhanced knowledge. Given the tabular data, $T$, the prototype, $P$, and the reference text, $R$, the learning objective of the sequence generator is the cross-entropy loss, deﬁned as Equation~\ref{ft}:
\begin{equation}
    L_{LM} = - \sum_{i=1}^{|R|} \log{P_{\mathscr{D+A}}}{(R_{i}|R_{<i}; \mathscr{E+A}([P:T]))} ,
\label{ft}
\end{equation}

where $\mathscr{E+A}$ and $\mathscr{D+A}$ denote the configurations where the KA is integrated subsequent to the encoder and decoder, respectively. While the proposed P2G framework is agnostic to the choice of the particular PLM, we leave such validation for future work.
\section{Experiments}

\begin{table}[htbp]
\caption{Overview of the data in Humans, Books, and Songs. Cell refers to the mean total cell count across the data sets. Essential Cell signifies the cells whose content is integral to the descriptions, indicating the relevance of specific data points to the narrative context. Table Length indicates the average word count for each table, providing insights into the data set's complexity. Corpus Length reflects the average word count of each prototype retrieved through the Information Retrieval (IR) system, highlighting the extent of background information considered. Description Length specifies the average word count of each description associated with the tables, illustrating the detail and breadth of the narrative content.}\protect
\begin{center}
\resizebox{0.85\columnwidth}{!}{

\begin{tabular}{lccc}
\toprule[0.25pt]
Statics            & Humans  & Books   & Songs   \\ \hline
Cell               & \textbf{12.85}  & 11.28  & 10.32  \\
Essential Cell     & \textbf{5.55}  & 4.78  & 4.64  \\
Table Length       & \textbf{392.26} & 288.51 & 291.18 \\
Corpus Length      & \textbf{21.29}  & 18.07  & 20.05  \\
Description Length & \textbf{25.60}  & 19.12  & 19.22  \\ \bottomrule[0.25pt]
\end{tabular}

}

\end{center}
\label{dataset-des}
\end{table}

\begin{table}[htbp]
\caption{GPT-3.5 turbo Results (zero-shot).}
\begin{center}
\resizebox{0.8\columnwidth}{!}{
\begin{tabular}{lccc}
\toprule[0.25pt]
      & ROUGE-4 & BLEU & PARENT-F \\ \hline
Book  & 4.17    & 1.26 & 8.19     \\
Human & 2.92    & 1.86 & 9.69     \\
Song  & 3.87    & 0.57 & 6.22     \\ 
 \bottomrule[0.25pt]  
\end{tabular}}
\end{center}
\label{gpt-3.5}
\end{table}

\begin{table*}[htbp]
\caption{BLEU-4 results for the Humans, Books, and Songs domains. All (R) are copied from the original paper. Bold denotes the best performance of this evaluation. The second-best ones are labeled with $\dag$}\protect
\begin{center}
\resizebox{\textwidth}{!}{
\begin{tabular}{lcccccccccccc}
\toprule[0.25pt]
Domain                      & \multicolumn{4}{c}{\textbf{Humans}} & \multicolumn{4}{c}{\textbf{Books}} & \multicolumn{4}{c}{\textbf{Songs}} \\ \cmidrule(r){1-1} \cmidrule(r){2-5} \cmidrule(r){6-9} \cmidrule(r){10-13}
Training set size           & 50     & 100     & 200     & 500    & 50     & 100     & 200    & 500    & 50     & 100     & 200    & 500    \\ \cmidrule(r){1-1} \cmidrule(r){2-5} \cmidrule(r){6-9} \cmidrule(r){10-13}
Structure-Aware (R)             & 2.9& 5.1 & 6.1 & 8.3& 7.3 & 6.8 &    7.8 &8.8 & 10.4 &12.0 &11.6 &13.1  \\
Pivot (R)             &14.9 &18.7& 25.3 &29.8& 23.1 &24.9&  27.0 & 29.8 & 26.2 &28.0 &29.2 & 31.7\\ \hline
SwitchPolicy (R)             &25.7&29.5&36.1&41.7& 34.3&36.2& 37.9& 40.3 & 36.1&37.2 & 39.4 &42.2 \\
TableGPT (R)                 & 29.8 & 34.5& 40.6 & 45.6& 35.1& 37.3& 38.5 &41.6&36.7& 37.8 & 39.3 & 42.3 \\
Prefix-Tuning+T5 (R)               &34.5 & 39.9 & 41.6  & 44.1 & 35.5 &  37.3 & 39.6 & 41.2 & 37.5 & 38.5 & 40.0 & 41.1 \\ 
BART-large               &39.2& 44.0&46.5& 49.6& 36.0& 37.2& 42.4& 45.6& 41.4& 43.7& 44.0&45.9\\ 
T5-prefix (R)             & 32.6& 37.1&41.7& 46.3& 34.2& 38.3&39.4& 42.9 &37.6 & 38.7 & 40.0& 43.5\\
PCG (R) &39.9&43.3&45.8&49.4&36.6 &36.9&39.0&45.6&38.0&41.7&42.5&44.5\\
AMG (R)                       &    -    &     -   &     -    &    49.0    &    -    &     -   &    -  &    43.9  &   -    &    -  &    -   &	45.1\\\hline
Retri-Gen(R)               & 7.4 & 10.3 & 13.2&16.5&12.1 &13.2&14.7& 15.9&13.4&14.3& 16.2&17.7\\
P2G (T5)(R)                 & 39.3& 42.6& $46.2^{\dag}$&$50.1^{\dag}$ &$41.2^{\dag}$&43.4&46.4& 49.2&42.8&45.9& 47.6&50.7\\
P2G (BART-large)         &$40.4^{\dag}$ & $44.5^{\dag}$ &46.0&49.9& 41.0&$44.4^{\dag}$&$47.6^{\dag}$& $49.8^{\dag}$ & $50.1^{\dag}$ & $50.4^{\dag}$&$52.2^{\dag}$& $53.1^{\dag}$\\\hline
AKG (BART-large) [\textbf{\textit{Ours}}]       &\textbf{43.3}& \textbf{45.9}  & \textbf{47.4} &\textbf{51.3} & \textbf{43.6} &\textbf{45.2}&\textbf{49.3}&\textbf{50.8}&\textbf{51.3}&\textbf{52.1}&\textbf{52.6}&\textbf{53.9}\\ \bottomrule[0.25pt]    
\end{tabular}}
\end{center}
\label{table-bleu}
\end{table*}

\begin{table*}[htbp]
\caption{ROUGE-4 results for the Humans, Books, and Songs domains. All (R) are copied from the original paper. The baseline, which does not evaluate the ROUGE-4 metric, is not shown in this table. Bold denotes the best performance of this evaluation. The second-best ones are labeled with $\dag$. The result of Humans with 200 training instances, AKG (BART-large) is slightly poorer than P2G (T5) but better than P2G (BART-large). Our methods still achieves the best performance in the case of similar parametric models}\protect
\begin{center}
\resizebox{\textwidth}{!}{
\begin{tabular}{lcccccccccccc}
\toprule[0.25pt]
Domain                      & \multicolumn{4}{c}{\textbf{Humans}} & \multicolumn{4}{c}{\textbf{Books}} & \multicolumn{4}{c}{\textbf{Songs}} \\ \cmidrule(r){1-1} \cmidrule(r){2-5} \cmidrule(r){6-9} \cmidrule(r){10-13}
Training set size           & 50     & 100     & 200     & 500    & 50     & 100     & 200    & 500    & 50     & 100     & 200    & 500    \\ \cmidrule(r){1-1} \cmidrule(r){2-5} \cmidrule(r){6-9} \cmidrule(r){10-13}
Structure-Aware (R)             & 0.1& 0.4 & 0.8 & 1.5 & 1.7 &1.5 &2.1 &2.4 & 4.1 &5.1 &4.7 &5.8\\
Pivot(R)             &3.2 &6.9& 14.1 &17.3 & 10.7 &13.3&  15.2 & 18.1 & 14.7 &16.2 &17.7 & 20.0\\ \hline
SwitchPolicy (R)             &14.1&16.2&22.1&28.3&22.5&23.1&25.0&27.6&26.2&28.6&30.1& 32.6\\
TableGPT (R)                 & 16.3 & 20.6 & 27.6 &32.4&24.0 &25.4 &26.7 &28.9 &27.1 & 29.4 &30.6 & 32.8 \\
BART-large               &23.7 &28.6&32.6&36.7&24.2&26.3&27.6&30.6 &30.8&32.5&33.1&35.2\\ 
T5-prefix (R)        &20.7&23.1 &28.8 & 33.2 & 21.2& 26.7&27.6&30.0 &28.1 & 29.2 &30.3 &33.9\\ \hline
Retri-Gen (R)                 & 0.7 &1.6 &2.7&4.1 &1.8 &2.0&2.4 &3.3 &2.7&3.1&4.3&4.9\\
P2G (T5)(R)                 &$27.9^{\dag}$ & $30.8^{\dag}$ &\textbf{34.0}&$37.3^{\dag}$&$28.3^{\dag}$ &$30.5^{\dag}$&$33.8^{\dag}$&$36.1^{\dag}$&33.0&35.7&37.5&40.1\\
P2G (BART-large)                &24.8&28.9&31.4&36.6&27.1&30.4&33.6&34.8&$42.4^{\dag}$&$41.6^{\dag}$&$43.7^{\dag}$&$44.3^{\dag}$\\\hline
AKG (BART-large) [\textbf{\textit{Ours}}]         &\textbf{28.2}&\textbf{31.5}&$33.3^{\dag}$&\textbf{39.2}&\textbf{29.6}&\textbf{31.1}&\textbf{34.6}&\textbf{36.9} &\textbf{43.7}&\textbf{43.9}&\textbf{44.8}&\textbf{46.1}\\ \bottomrule[0.25pt]    
\end{tabular}}
\end{center}
\label{table-rouge}
\end{table*}

\subsection{Data Sets and Experimental Setup}
Adhering to the experimental framework established by \cite{chen2020few}, we conduct an evaluation of our method across three distinct benchmarks within the WIKIBIO data set \cite{liu2018table}, specifically: Humans, Books, and Songs. Table~\ref{dataset-des} furnishes a comprehensive overview of the data sets spanning these domains. Compared to the Books and Songs data sets, the Humans data set is characterized by statistical attributes. Notably, the statistical attributes Cell, Essential Cell, and Table Length reveal that the tabular data within the Humans domain encapsulates a more significant amount of critical information pertinent to their respective descriptions. We perform experiments in few-shot settings by varying the training set size from $\left\{50,100,200,500\right\}$. The size of the validation set is 1000, and the size of the test set is 13587, 5252, and 11879 for Humans, Books, and Songs, respectively.\newline

To elucidate the intricacies of our experimental task, as depicted in Table~\ref{gpt-3.5}, we undertook a supplementary analysis focusing on the performance metrics of the extensively recognized, large-scale GPT-3.5 turbo model in a zero-shot context \cite{ye2023comprehensive}. This examination aimed to contextualize our results amidst the latest developments in the field of generative pre-trained models. Our analysis underscores that despite the GPT-3.5 model's prevalent success across many applications, it encounters challenges in synthesizing coherent narratives from linearized tabular data. \newline

We leverage the design and settings of \cite{su2021few} for the development of the Information Retrieval (IR) system. Lucene \cite{yang2017anserini}, \cite{bialecki2012apache} was utilized to pre-index all sentences contained within the English Wikipedia corpus as of the December 2018 dump. This pre-indexing process enables the IR system to retrieve a set of 100 candidate sentences efficiently, referred to as candidates $B$, for each table under consideration. Following this retrieval, the role of the prototype selector emerges, with its primary function being to discern and in alignment with the approach established by \cite{su2021few}, we select the top three sentences from the candidate set R, hereby designated as the prototypes. In calibrating the prototype selector during its training phase, we assign the value of 5 to the parameter k, as specified in Equation~\ref{prototype-object}. This procedural framework assures a methodical selection strategy to identify the sentences most pertinent to act as prototypes, optimizing the subsequent processing phases. In the course of developing the knowledge adapter, we employed an unlabeled corpus sourced from Wikipedia, as detailed by \cite{su2021few}. Our experimental framework across three distinct domains was characterized by a deliberate modification of the training data set size, which was established at 500 instances. Conversely, the composition of the test set was defined by a total of 1000 instances. \newline

\begin{table*}[htbp]
\caption{PARENT-F results for the Humans, Books, and Songs domains. All (R) are copied from the original paper. The baseline, which does not evaluate the PARENT-F metric, is not shown in this table. Bold denotes the best performance of this evaluation. The second-best ones are labeled with $\dag$}\protect
\begin{center}
\resizebox{\textwidth}{!}{
\begin{tabular}{lcccccccccccc}
\toprule[0.25pt]
Domain                      & \multicolumn{4}{c}{\textbf{Humans}} & \multicolumn{4}{c}{\textbf{Books}} & \multicolumn{4}{c}{\textbf{Songs}} \\ \cmidrule(r){1-1} \cmidrule(r){2-5} \cmidrule(r){6-9} \cmidrule(r){10-13}
Training set size           & 50     & 100     & 200     & 500    & 50     & 100     & 200    & 500    & 50     & 100     & 200    & 500    \\ \cmidrule(r){1-1} \cmidrule(r){2-5} \cmidrule(r){6-9} \cmidrule(r){10-13}
SwitchPolicy (R)        & 30.6 & 34.6 & 40.5 & 45.6 & 42.7 & 42.8 & 43.4 & 44.9 & 40.2 & 41.7 & 44.0 & 44.8 \\
BART-large                & 44.4 & 48.2 & 50.0 & 51.3 & 42.9 & 45.5 & 46.3 & 47.3 & 44.7 & 46.8 & 44.9  & 47.1 \\
AMG (R)                        & 43.6 &  47.7 & 50.1 & $51.9^{\dag}$ & 43.4 & 46.0 & 47.5 & 48.6 & 42.0 & 43.3 & 45.9 & 46.9 \\
Preﬁx-Tuning+T5 (R)        & 39.3  & 40.6 & 41.8 & 42.1 & 32.8 & 34.8 & 36.0 & 36.8 & 34.4 & 36.1 & 36.0 & 34.6 \\
PCG (R) & $46.7^{\dag}$  & $48.3^{\dag}$ & \textbf{50.4} & 51.8 & $46.3^{\dag}$ & 46.2 & 47.5 & 49.3 & 44.8 & 45.7 & 46.9 & 46.0 \\ \hline
P2G (BART-large)     & 44.3  & 48.2 & 49.3 & 51.4 & 44.4 & $47.4^{\dag}$ & \textbf{49.2} & $49.7^{\dag}$ & $45.5^{\dag}$ & $47.0^{\dag}$ & \textbf{48.5} & $47.9^{\dag}$ \\ \hline
AKG (BART-large) [\textbf{\textit{Ours}}]                              & \textbf{47.1} & \textbf{49.8} & $50.3^{\dag}$  & \textbf{52.6} & \textbf{46.9} & \textbf{49.1} & $48.7^{\dag}$  & \textbf{49.9} & \textbf{47.9} & \textbf{48.0} & $48.3^{\dag}$ & \textbf{49.3}  \\ \bottomrule[0.25pt]  
\end{tabular}}
\end{center}
\label{table2}
\end{table*}

In our study, the BART-large model \cite{lewis2020bart} serves as the foundational generator, implemented via the Hugging Face Library \cite{wolf2020transformers}. To ensure optimal performance across all pretraining tasks, we employ a learning rate of $3 \times 10^{-5}$, utilizing the Adam optimization algorithm \cite{DBLP:journals/corr/KingmaB14}. This computational framework is executed on an NVIDIA GeForce RTX 3090 GPU, providing the necessary hardware acceleration to efficiently manage the extensive computational demands associated with processing large-scale data sets and model training.\newline

\subsection{Baseline Models}
In our study, we juxtapose our approach against preceding state-of-the-art methodologies in the domain of few-shot table-to-text generation, which are utilized as benchmark baselines. These baseline methodologies are categorized into three distinct groups for comparative analysis: naive sequence-to-sequence (seq2seq)-based, PLM-based, and retrieval-based methods. This classification facilitates a comprehensive evaluation, allowing us to delineate our approach's advancements over a diverse array of existing strategies within the scope of few-shot table-to-text generation.\newline

\noindent \textbf{Naive seq2seq-based models}:
\begin{itemize}
\item Structure-Aware \cite{liu2018table}: a structure-aware seq2seq architecture consisting of a field-gating encoder and a dual-attention-based generator, can generate coherent and fluent descriptions.
\item Pivot \cite{ma2019key}: a two-stage generation model, consisting of key fact prediction from tables and surface realization for generation, achieves remarkable performance for low-resource table-to-next generation.
\end{itemize}

\noindent \textbf{PLM-based models}:
\begin{itemize}
\item Switch Policy with PLM \cite{chen2020few}: the ﬁrst approach suggested for the few-shot NLG task based on PLM. The authors suggest a switch policy to balance generating or copying from the table contents. Switch+GPT2 and Switch+BART-large were implemented by \cite{chen2020few}, and \cite{luo2022few}, respectively.
\item TableGPT \cite{gong2020tablegpt}: a further study based on Swtich+GPT2 that utilizes two auxiliary tasks for structure construction and content matching to generate faithful text.
\item BART-Large \cite{lewis2020bart}: a powerful PLM for generative challenges.
\item T5-Prefix \cite{raffel2020exploring}: a T5 PLM, which is utilized for conditional generation with special prefix tokens.
\item AMG \cite{zhao-etal-2021-attend-memorize}: a PLM-based approach with multigrain attention on table slots and tokens with a dynamic memory mechanism to backtrack the allocation of table slots.
\item Prefix-tuning \cite{li2021prefix}: a prompt-tuning method that prepends a continuous token to preserve prior knowledge of the PLM. The performance of few-shot table-to-text generation was explored by \cite{luo2022few}.
\item PCG \cite{luo2022few}: a prompt-tuning method with both prefix-tuning and hard prompt to control generation content.
\end{itemize}

\noindent \textbf{Retrieval-based models}:
\begin{itemize}
\item Retri-Gen \cite{wu2019response}: a retrieval-based approach that retrieves and edits a prototype response from a predefined index for sentence generation.
\item P2G \cite{su2021few}: the authors proposed a retrieval-based framework that utilizes an IR system to provide a prototype for improving generation quality.
\end{itemize}

\subsection{Automatic Evaluation}
Following the previous settings \cite{chen2020few, su2021few, luo2022few}, we perform automatic evaluation with BLEU-4 \cite{papineni2002bleu} and ROUGE-4 \cite{lin2004rouge} to measure the similarity between the generation of systems and the reference descriptions. BLEU-4 calculates the geometric mean of the precision over 4 grams of the output text. ROUGE-4 counts the number of overlapping 4 gram tokens between the generated description and the ideal summaries. In addition, we investigate the automatic evaluation of PARENT \cite{dhingra2019handling}, a token overlap-based metric that estimates the ﬁdelity of the generated sentence with respect to both the original table and the reference sentence. In our experiments, we report the F1 score of PARENT, denoted PARENT-F. \newline

\begin{table}[htbp]
\caption{Human evaluation results. $\uparrow$ denotes the higher the better and $\downarrow$ denotes the lower the better. Bold denotes the best performance of this evaluation}
\begin{center}
\resizebox{\linewidth}{!}{
\begin{tabular}{lccc}
\toprule[0.25pt]
                       & \# Sup $\uparrow$ & \# Cont $\downarrow$  & Fluency $\uparrow$  \\ \hline
P2G (BART-large)         & 3.99   & 0.78    & 2.35     \\
AKG (BART-large) [\textbf{\textit{Ours}}] & \textbf{4.20}   & \textbf{0.56}    & \textbf{2.74}     \\
 \bottomrule[0.25pt]  
\end{tabular}}
\end{center}
\label{table3}
\end{table}

Table~\ref{table-bleu} and Table~\ref{table-rouge} show the BLEU4 and ROUGE-4 results of our experiments. Our approach achieves state-of-the-art performance in the three domains, demonstrating the robustness and universality of our approach. Under near-parametric conditions, our approach provides a significant boost compared to previous methods. As shown in Fig.~\ref{Figure4}, compared with P2G (BART-large), which produces the second-best result with a similar number of parameters, our approach improves by, on average, 4\%, 3\%, and 2\% for BLEU and 9\%, 5\%, and 4\% for ROUGE on the Humans, Books, and Songs data sets, respectively. The results show that our method can produce fluent descriptions. \newline

\begin{table*}[htbp]
\caption{Ablation study of the BLEU4 results for the Humans, Books, and Songs domains. Bold denotes the best performance of this evaluation}
\begin{center}
\resizebox{1.7\columnwidth}{!}{
\begin{tabular}{lcccccccccccc}
\toprule[0.25pt]
Domain                      & \multicolumn{4}{c}{\textbf{Humans}} & \multicolumn{4}{c}{\textbf{Books}} & \multicolumn{4}{c}{\textbf{Songs}} \\ \cmidrule(r){1-1} \cmidrule(r){2-5} \cmidrule(r){6-9} \cmidrule(r){10-13}
Training set size           & 50     & 100     & 200     & 500    & 50     & 100     & 200    & 500    & 50     & 100     & 200    & 500    \\ \cmidrule(r){1-1} \cmidrule(r){2-5} \cmidrule(r){6-9} \cmidrule(r){10-13}
AKG        & \textbf{43.3}  & \textbf{45.9} & 47.4& \textbf{51.3} & \textbf{43.6} &\textbf{45.2} &\textbf{49.3}& \textbf{50.8} & \textbf{51.3} & \textbf{52.1} & \textbf{52.6}&\textbf{53.9} \\
-KA                & 40.4 & 44.5& 46.0 & 49.9& 41.0 & 44.4 &47.6 & 49.8&50.1 & 50.4 & 52.2 & 53.1\\
-PT              & 40.7& 44.8 & \textbf{48.0} & 50.9& 36.4& 37.1 & 42.6& 45.9& 41.4& 43.4 &44.6 & 46.1\\
-KA\&PT                       & 39.2 &  44.0 & 46.5 & 49.6&36.0 & 37.2 & 42.4 & 45.6& 41.4& 43.7& 44.0 & 45.9
\\ \bottomrule[0.25pt]
\end{tabular}}
\end{center}
\label{table-ableu}

\end{table*}

Concerning the results of BLEU4 and ROUGE-4, the PLM-based methods significantly improve the fluency and coherence of the yielded sentences compared to the naive Seq2seq methods. By extending GPT2 \cite{radford2019language}, TableGPT \cite{gong2020tablegpt}, and SwitchPolicy, \cite{chen2020few} achieved remarkable performance over the previous naive method. By increasing the number of model parameters and optimizing the encoder--decoder, T5 \cite{raffel2020exploring}, BART-large \cite{lewis2020bart}, Prefix-Tuning+T5 \cite{li2021prefix}, and PCG, \cite{luo2022few} further improved the generation quality. However, the lack of domain-specific knowledge of PLMs becomes a bottleneck to bridging the gap between tabular data and descriptions. The P2G research of \cite{su2021few} introduced a retrieval-based method via the unlabeled domain-specific knowledge corpus and provided a new way to overcome the shortcomings of PLM-based methods. However, this method leaves out most of the information. Our approach provides an effective solution that targets this shortcoming according to the results. \newline

\begin{table*}[htbp]
\caption{Ablation study of ROUGE-4 results for the Humans, Books, and Songs domains. Bold denotes the best performance of this evaluation}
\begin{center}
\resizebox{1.7\columnwidth}{!}{
\begin{tabular}{lcccccccccccc}
\toprule[0.25pt]
Domain                      & \multicolumn{4}{c}{\textbf{Humans}} & \multicolumn{4}{c}{\textbf{Books}} & \multicolumn{4}{c}{\textbf{Songs}} \\ \cmidrule(r){1-1} \cmidrule(r){2-5} \cmidrule(r){6-9} \cmidrule(r){10-13}
Training set size           & 50     & 100     & 200     & 500    & 50     & 100     & 200    & 500    & 50     & 100     & 200    & 500    \\ \cmidrule(r){1-1} \cmidrule(r){2-5} \cmidrule(r){6-9} \cmidrule(r){10-13}
AKG        & \textbf{28.2}  & \textbf{31.5} & 33.3 & \textbf{39.2} & \textbf{29.6} &\textbf{31.1} &\textbf{34.6}& \textbf{36.9} & \textbf{43.7} & \textbf{43.9} & \textbf{44.8}&\textbf{46.1} \\
-KA                & 24.8 & 28.9 & 31.4 & 36.6 & 27.1 & 30.4 &33.6 & 34.8 &42.4 &41.6 & 43.7 & 44.3 \\
-PT               & 26.2 & 30.9 & \textbf{34.6} & 37.4 & 24.1 & 26.4 & 28.9& 31.7 & 30.8 & 33.2 &34.3  & 37.4 \\
-KA\&PT                        & 23.7 & 28.6 & 32.6 & 36.7 &24.2 & 26.3 & 27.6 & 30.6 & 30.8 & 32.5 & 33.1 & 35.2 
\\ \bottomrule[0.25pt]  
\end{tabular}}
\end{center}
\label{table-arouge}
\end{table*}

\begin{table*}[htbp]
\caption{Ablation study of the PARENT-F results for the Humans, Books, and Songs domains. Bold denotes the best performance of this evaluation}
\begin{center}
\resizebox{1.7\columnwidth}{!}{
\begin{tabular}{lcccccccccccc}
\toprule[0.25pt]
Domain                      & \multicolumn{4}{c}{\textbf{Humans}} & \multicolumn{4}{c}{\textbf{Books}} & \multicolumn{4}{c}{\textbf{Songs}} \\ \cmidrule(r){1-1} \cmidrule(r){2-5} \cmidrule(r){6-9} \cmidrule(r){10-13}
Training set size           & 50     & 100     & 200     & 500    & 50     & 100     & 200    & 500    & 50     & 100     & 200    & 500    \\ \cmidrule(r){1-1} \cmidrule(r){2-5} \cmidrule(r){6-9} \cmidrule(r){10-13}
AKG        			  & \textbf{47.1} & \textbf{49.8}  & 50.3  & 52.6  & \textbf{46.9}  & \textbf{49.1}  & 48.7  & \textbf{49.9}  & \textbf{47.9} &\textbf{48.0}  & 48.3  & \textbf{49.3}  \\
-KA          			  & 44.3  & 48.2  & 49.3  & 51.4  & 44.4 & 47.4  & \textbf{49.2} &49.7  & 45.5  & 47.0  &\textbf{48.5} & 47.9  \\
-PT          			   & 46.4  & 49.5  & \textbf{51.0}  & \textbf{53.0}  & 45.1  & 45.8  & 47.4  & 45.9  & 45.5  & 46.8  &46.7& 47.3  \\
-KA\&PT             &44.4  & 48.2 &  49.5  & 51.3  &42.9 & 45.5  & 46.3  & 47.3  & 44.7  & 46.8  &44.9  & 47.1 
\\ \bottomrule[0.25pt]  
\end{tabular}}
\end{center}
\label{table5}
\end{table*}

Our approach also achieves better performance for PARENT-F than the other baseline methods. According to the results, compared to P2G (BART-large), our performance is better by, on average, 1.7\% for nine terms and, on average, only 0.5\% worse for three terms in PARENT-F. As we use the result of ROUGE-4 and BLEU as the primary evaluation standards throughout our training process, partial PARENT-F accuracy is sacrificed. At the same time, during the human evaluation, we found that the knowledge-augmentation method also affects the PARENT-F score while enriching the generated content.

\subsection{Human Evaluation}
We also conduct a human evaluation to compare the AKG with the closest baseline, P2G (BART-large). All volunteers are postgraduate computer science students. Following the settings in \cite{chen2020few}, we evaluate each generated sentence with two tasks: faithfulness and fluency evaluation. The experiments are performed on the Humans data set with 100 training instances. We randomly select 100 generated sentences with the corresponding tabular data. In order to reduce variance caused by the participants, each example is scored by three different people. \newline

Faithfulness aims to evaluate the correct information in the generated sentences. Only all information supported by the table makes the generated sentence faithful. Throughout the evaluation, each evaluator was asked to count the number of contained facts supported by the table data, noted as $\#Sup$, and the number of contradictory facts, noted as $\#Cont$. We report the average number of $\#Sup$ and $\#Cont$ in Table~\ref{table3}. Fluency tries to evaluate the fluency of the generated sentences. A sentence is fluent if it is grammatically correct and natural. The raters were asked to rate the output in terms of ﬂuency on a three-point Likert scale (0, 1, or 2). We report the average results in Table~\ref{table3}. The results show that our method provides a significant improvement over P2G (BART-large) for all metrics (sign test with a p-value \textless 0.05).

\subsection{Ablation Study}
We conduct an ablation study on the BART-large model to systematically assess the impact of each proposed technique. In this context, the suffix ``-KA" signifies the exclusion of knowledge adapter, thereby relying solely on PT, which aligns with the process of training P2G utilizing BART-large. Conversely, ``-PT" denotes the removal of the prototype selector, entailing the application of KA for knowledge injection and leveraging the BART-large model as the primary generator to craft fluent descriptions from linearized tabular data. Additionally, ``-KA\&PT" illustrates the scenario where only the backbone generator (BART-large) is employed, devoid of any augmentations. Table~\ref{table-ableu}, Table~\ref{table-arouge}, and Table~\ref{table5} demonstrate the results of BLEU4, ROUGE-4, and PARENT-F, respectively, for the Humans, Books, and Songs domains, and for training set sizes of 50, 100, 200, and 500. \newline

We further investigate the impact of the number of prototypes, $n$ in Equation~\ref{prototype-object}, on model performance with the setting of the suffix ``-KA". Specifically, we train both BART-large and BART-small models using 50 and 500 instances from the Humans data set while varying the size of $n$. As illustrated in Table~\ref{bart-comparison}, the outcomes encompass metrics such as BLEU, ROUGE-4, and PARENT-F. For the Bart model, our findings indicate that when $n$ is relatively small, the performance metrics remain consistent. However, as $n$ approaches 10, a noticeable decline in performance is observed. Contrarily, the Bart-large model exhibits negligible sensitivity to variations in the number of prototypes, underscoring a distinct behavior in comparison to its smaller counterpart. This observed divergence is presumably due to the incremental inclusion of $n$ introducing ancillary or non-pertinent information within the tabular context, thus infusing noise and detrimentally impacting model performance. However, this adverse effect is conspicuously attenuated in models characterized by a broader parameter set and an increased volume of training instances, thereby indicating a diminished vulnerability to noise-induced degradation. Additionally, an evaluative comparison between models integrated with the PT module and those devoid of it—irrespective of employing BART-small or BART-large—unveils a tangible enhancement attributable to the PT module. Aligning with the experimental framework proposed in \cite{su2021plan}, we adopted $n=3$ in our experimental setup to maintain consistency and relevance in our investigation. \newline

We also explored the impact of utilizing an unlabeled corpus on the performance of KA, particularly in the context of the ``-PT" suffix. Following the methodology outlined in previous work \cite{su2021few}, we divided the unlabeled corpus sourced from Wikipedia into two segments: 500 sentences for training and 1,000 sentences for testing purposes. In this experimental setup, we trained the BART-large model using 50 instances from the Humans data set while systematically varying the size of the unlabeled corpus in increments of 100, from 100 to 500 sentences. The results, detailed in Table~\ref{corpus-limitation}, were evaluated using metrics such as BLEU, ROUGE-4, and PARENT-F. Our findings suggest that there is a notable improvement in performance with the increase in the size of the unlabeled corpus used. However, after the data volume exceeds 400 sentences, the experimental results begin to decline slightly due to the introduction of noise adversely affecting the model. Nonetheless, when compared to models without KA integration, a significant enhancement in overall results is observed, underscoring the substantial improvement brought about by incorporating KA.\newline

The outcomes of this ablation study elucidate that integrating both the PT and the KA substantially enhances performance beyond that achievable by the backbone generator alone, with the exception observed in the Humans data set. An examination of the data sets, as detailed in Table~\ref{dataset-des}, reveals that the tabular data about Humans encompasses more information than the other two data sets. Paradoxically, this increased information content reduces complexity yet simultaneously diminishes the performance efficacy of PT. The comprehensive application of all proposed techniques results in further amelioration of experimental outcomes, underscoring the synergistic potential of these methodologies in enhancing model performance across diverse data contexts.

\begin{table*}[htbp]
\caption{An empirical investigation was undertaken to ascertain the influence of prototype quantity ($n$), employing the Human data set under both 50-shot and 500-shot training scenarios.}
\begin{center}
\resizebox{2\columnwidth}{!}{
\begin{tabular}{lcccccccccccc}
\toprule[0.25pt]
\multicolumn{6}{c}{BART-small} & \multicolumn{6}{c}{BART-large} \\ \cmidrule(r){2-7} \cmidrule(r){8-13}
n & \multicolumn{3}{c}{\textbf{50}} & \multicolumn{3}{c}{\textbf{500}} & \multicolumn{3}{c}{\textbf{50}} & \multicolumn{3}{c}{\textbf{500}}\\ 
\cmidrule(r){2-4} \cmidrule(r){5-7} \cmidrule(r){8-10} \cmidrule(r){11-13}
 &BLEU & ROUGE-4 & PARENT-F & BLEU & ROUGE-4 & PARENT-F & BLEU & ROUGE-4 & PARENT-F & BLEU & ROUGE-4 & PARENT-F \\ 
\cmidrule(r){1-1} \cmidrule(r){2-4} \cmidrule(r){5-7} \cmidrule(r){8-10} \cmidrule(r){11-13} 
Nah & $\mathbf{38.9}$ & 24.3 & 44.9 & 49.4 & 38.4 & 51.6 & 39.2 & 23.7 & 44.4 & 49.6 & 36.7 & 51.3 \\
1 & 37.6 & 25.1 & 43.7 & 49.0 & 37.5 & 52.3 & 39.8 & 25.9 & 44.6 & 50.2 & 37.2 & 52.0 \\
2 & 38.0 & 25.5 & $\mathbf{45.0}$ & 49.2 & 37.8 & 52.4 & 40.0 & 23.4 & 44.1 & 51.2 & 37.1 & $\mathbf{53.3}$ \\
3 & 38.3 & $\mathbf{25.5}$ & 44.6 & $\mathbf{50.3}$ & 37.9 & $\mathbf{53.4}$ & 40.4 & 24.8 & 44.3 & 49.9 & 36.6 & 51.4 \\
4 & 38.2 & 25.1 & 44.4 & 49.7 & $\mathbf{38.6}$ & 53.0 & $\mathbf{41.4}$ & 25.2 & 45.2 & $\mathbf{50.6}$ & 37.1 & 52.0 \\
5 & 38.1 & 25.1 & 43.8 & 49.4 & 38.3 & 52.9 & 41.0 & 25.8 & 44.7 & $\mathbf{50.6}$ & $\mathbf{37.6}$ & 52.3 \\
6 & 38.1 & 25.0 & 43.4 & 49.1 & 37.9 & 52.3 & 40.3 & $\mathbf{26.3}$ & 44.3 & 50.3 & 35.8 & 52.4 \\
7 & 38.1 & 23.7 & 43.4 & 48.4 & 37.8 & 52.3 & 41.0 & 24.9 & 45.1 & 50.5 & 36.3 & 51.7 \\
8 & 38.0 & 25.0 & 44.0 & 49.3 & 37.9 & 52.8 & 41.2 & 25.6 & 45.0 & 50.1 & 36.1 & 51.9 \\
9 & 36.9 & 24.3 & 42.0 & 48.8 & 37.5 & 52.2 & 41.3 & 25.1 & $\mathbf{45.9}$ & 50.0 & 36.0 & 51.2 \\
10 & 37.1 & 24.6 & 41.5 & 48.0 & 37.0 & 51.7 & 40.2 & 25.1 & 44.6 & 50.4 & 36.2 & 52.7 \\
\bottomrule[0.25pt]  
\end{tabular}}
\end{center}
\label{bart-comparison}
\end{table*}

\begin{table}[htbp]
\caption{An empirical investigation was undertaken to ascertain the influence of utilizing the external unlabeled corpus sentences, employing the Human data set under both 50-shot training scenarios.}
\begin{center}
\resizebox{0.8\columnwidth}{!}{
\begin{tabular}{lccc}
\toprule[0.25pt]
Utilized Corpus & BLEU & ROUGE-4 & PARENT-F \\ \hline
Nah              & 39.2      &   23.7      &   44.4       \\
100             & 40.0     &    25.3     &    45.7      \\
200             & $\mathbf{41.7}$ & 26.7    & 46.5     \\
300             & 41.6 & 26.1    & 46.6     \\
400             & 41.4 & $\mathbf{27.5}$    & $\mathbf{46.9}$     \\
500            & 40.7 & 26.2    & 46.4     \\
\bottomrule[0.25pt] 
\end{tabular}}
\end{center}
\label{corpus-limitation}
\end{table}

\subsection{Case Studies}
\begin{figure*}[!t]
\centering
\includegraphics[scale=0.38]{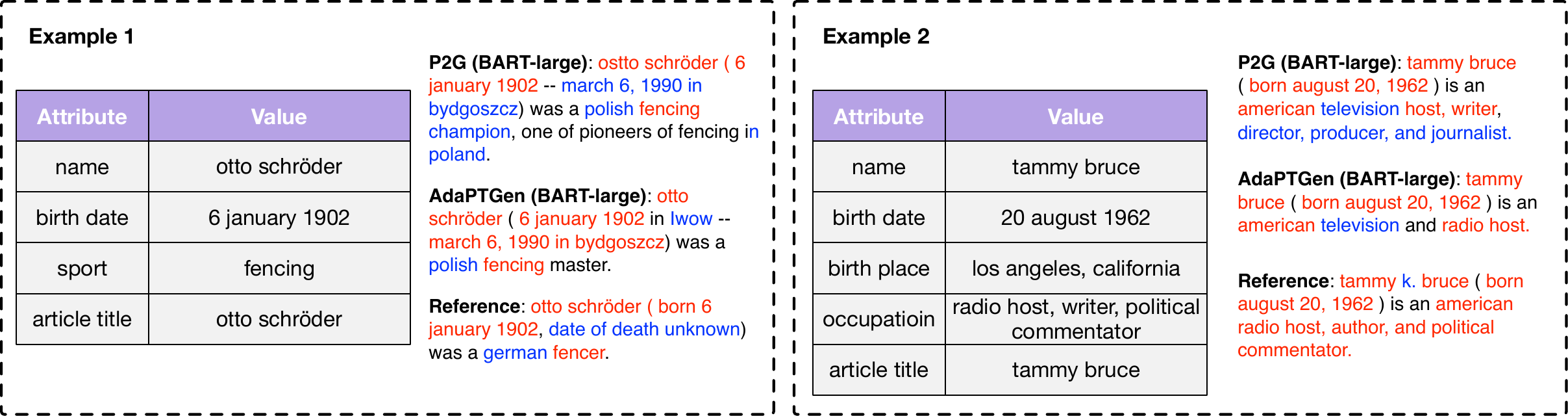}
\caption{Two example tables from the Human test data set, where the yielded texts from the different methods are trained with 100 training data points. The blue text denotes that the (incorrect) fact conflicts with the information in the original table. The red text indicates the fact is supported by the tabular data.}
\label{Figure4}
\end{figure*}

In Fig.~\ref{Figure4}, we present two generated examples of our model against the strongest baseline P2G (BART-large), along with the references from the Humans domain under 100 instances. The blue text indicates facts that are supported by the tabular data, and the red text indicates facts that are incorrect or not shown in the tabular data. \newline

As seen in the first example, all attribute-value pairs are mentioned in the references and the two generated sentences. The reference sentences refer to ``date of death unknown", which contradicts the original input. The two generated sentences introduce the correct ``date of death" and the corresponding place via the domain-specific corpus. However, the P2G (BART-large) framework yields the fragment ``fencing champion", which is far from the reference sentence and the tabular data. In contrast, the AKG framework performs better in balancing the use of the facts of the original table and the domain-specific knowledge. A similar issue can be seen in the second example. P2G (BART-large) generates the fragment ``director, producer, and journalist", which is not mentioned in the tabular data or the references. \newline

These results further illustrate that the AKG framework brings benefits in terms of balancing the use of facts from the original spreadsheet and domain-specific knowledge. The fluency and accuracy also improve.

\section{Conclusions}
In this paper, we propose the AKG framework for few-shot table-to-text generation. Taking advantage of the adapting prompt design and the modularized pretraining strategy, we inject representations of the linguistic and semantic patterns of table-related descriptions and exploit a large domain-specific knowledge corpus fully. With the modularization strategy, our framework can devise various pretraining tasks to enhance the generative task, achieving high fluency and accuracy. Experimental results on three benchmark data sets show that our framework achieves superior performance in both fluency and faithfulness metrics. Our code and other related resources can be found in \url{https://github.com/sjtugzx/AKG}.

\section{Acknowledgments} 
We thank the anonymous reviewers for their thoughtful comments. We thank Zhouhan Lin, Zhiyu Chen, and Junxian He for their valuable advice. Guanjie Zheng is the corresponding author of this paper.  We acknowledge the financial support of the Ministry of Science and technology of the People's Republic China grant \#2022YFB3904200 and National Science Foundation of China grant \#42050105, \#61960206002, \#62020106005, \#62032020, \#62061146002, \#62272301. This work is supported by the Deep-time Digital Earth (DDE) Big Science Program.

\bibliographystyle{IEEEtran}
\bibliography{main}

\begin{thebibliography}{10}
\providecommand{\url}[1]{#1}
\csname url@samestyle\endcsname
\providecommand{\newblock}{\relax}
\providecommand{\bibinfo}[2]{#2}
\providecommand{\BIBentrySTDinterwordspacing}{\spaceskip=0pt\relax}
\providecommand{\BIBentryALTinterwordstretchfactor}{4}
\providecommand{\BIBentryALTinterwordspacing}{\spaceskip=\fontdimen2\font plus
\BIBentryALTinterwordstretchfactor\fontdimen3\font minus
  \fontdimen4\font\relax}
\providecommand{\BIBforeignlanguage}[2]{{%
\expandafter\ifx\csname l@#1\endcsname\relax
\typeout{** WARNING: IEEEtran.bst: No hyphenation pattern has been}%
\typeout{** loaded for the language `#1'. Using the pattern for}%
\typeout{** the default language instead.}%
\else
\language=\csname l@#1\endcsname
\fi
#2}}
\providecommand{\BIBdecl}{\relax}
\BIBdecl

\bibitem{gatt2018survey}
A.~Gatt and E.~Krahmer, ``Survey of the state of the art in natural language
  generation: Core tasks, applications and evaluation,'' \emph{Journal of
  Artificial Intelligence Research}, vol.~61, pp. 65--170, 2018.

\bibitem{chen2021finqa}
Z.~Chen, W.~Chen, C.~Smiley, S.~Shah, I.~Borova, D.~Langdon, R.~Moussa,
  M.~Beane, T.-H. Huang, B.~R. Routledge \emph{et~al.}, ``Finqa: A dataset of
  numerical reasoning over financial data,'' in \emph{Proceedings of the 2021
  Conference on Empirical Methods in Natural Language Processing}, 2021, pp.
  3697--3711.

\bibitem{ghazvininejad2018knowledge}
M.~Ghazvininejad, C.~Brockett, M.-W. Chang, B.~Dolan, J.~Gao, W.-t. Yih, and
  M.~Galley, ``A knowledge-grounded neural conversation model,'' in
  \emph{Proceedings of the AAAI Conference on Artificial Intelligence},
  vol.~32, no.~1, 2018.

\bibitem{chen2020open}
W.~Chen, M.-W. Chang, E.~Schlinger, W.~Y. Wang, and W.~W. Cohen, ``Open
  question answering over tables and text,'' in \emph{International Conference
  on Learning Representations}, 2020.

\bibitem{he2017learning}
H.~He, A.~Balakrishnan, M.~Eric, and P.~Liang, ``Learning symmetric
  collaborative dialogue agents with dynamic knowledge graph embeddings,'' in
  \emph{Proceedings of the 55th Annual Meeting of the Association for
  Computational Linguistics (Volume 1: Long Papers)}, 2017, pp. 1766--1776.

\bibitem{wiseman2017challenges}
S.~Wiseman, S.~M. Shieber, and A.~M. Rush, ``Challenges in data-to-document
  generation,'' in \emph{Proceedings of the 2017 Conference on Empirical
  Methods in Natural Language Processing}, 2017, pp. 2253--2263.

\bibitem{murakami2017learning}
S.~Murakami, A.~Watanabe, A.~Miyazawa, K.~Goshima, T.~Yanase, H.~Takamura, and
  Y.~Miyao, ``Learning to generate market comments from stock prices,'' in
  \emph{Proceedings of the 55th Annual Meeting of the Association for
  Computational Linguistics (Volume 1: Long Papers)}, 2017, pp. 1374--1384.

\bibitem{hasan2019clinical}
S.~A. Hasan and O.~Farri, ``Clinical natural language processing with deep
  learning,'' in \emph{Data science for healthcare}.\hskip 1em plus 0.5em minus
  0.4em\relax Springer, 2019, pp. 147--171.

\bibitem{guo2023towards}
Z.~Guo, J.~Zhou, J.~Qi, M.~Yan, Z.~He, G.~Zheng, Z.~Lin, X.~Wang, and C.~Zhou,
  ``Towards controlled table-to-text generation with scientific reasoning,''
  \emph{arXiv preprint arXiv:2312.05402}, 2023.

\bibitem{lebret2016neural}
R.~Lebret, D.~Grangier, and M.~Auli, ``Neural text generation from structured
  data with application to the biography domain,'' in \emph{Proceedings of the
  2016 Conference on Empirical Methods in Natural Language Processing}, 2016,
  pp. 1203--1213.

\bibitem{iso2019learning}
H.~Iso, Y.~Uehara, T.~Ishigaki, H.~Noji, E.~Aramaki, I.~Kobayashi, Y.~Miyao,
  N.~Okazaki, and H.~Takamura, ``Learning to select, track, and generate for
  data-to-text,'' in \emph{Proceedings of the 57th Annual Meeting of the
  Association for Computational Linguistics}, 2019, pp. 2102--2113.

\bibitem{parikh2020totto}
A.~Parikh, X.~Wang, S.~Gehrmann, M.~Faruqui, B.~Dhingra, D.~Yang, and D.~Das,
  ``Totto: A controlled table-to-text generation dataset,'' in
  \emph{Proceedings of the 2020 Conference on Empirical Methods in Natural
  Language Processing (EMNLP)}, 2020, pp. 1173--1186.

\bibitem{chen2020few}
Z.~Chen, H.~Eavani, W.~Chen, Y.~Liu, and W.~Y. Wang, ``Few-shot nlg with
  pre-trained language model,'' in \emph{Proceedings of the 58th Annual Meeting
  of the Association for Computational Linguistics}, 2020, pp. 183--190.

\bibitem{gong2020tablegpt}
H.~Gong, Y.~Sun, X.~Feng, B.~Qin, W.~Bi, X.~Liu, and T.~Liu, ``Tablegpt:
  Few-shot table-to-text generation with table structure reconstruction and
  content matching,'' in \emph{Proceedings of the 28th International Conference
  on Computational Linguistics}, 2020, pp. 1978--1988.

\bibitem{li2021prefix}
X.~L. Li and P.~Liang, ``Prefix-tuning: Optimizing continuous prompts for
  generation,'' in \emph{Proceedings of the 59th Annual Meeting of the
  Association for Computational Linguistics and the 11th International Joint
  Conference on Natural Language Processing (Volume 1: Long Papers)}, 2021, pp.
  4582--4597.

\bibitem{luo2022few}
Y.~Luo, M.~Lu, G.~Liu, and S.~Wang, ``Few-shot table-to-text generation with
  prefix-controlled generator,'' in \emph{Proceedings of the 29th International
  Conference on Computational Linguistics}, 2022, pp. 6493--6504.

\bibitem{su2021few}
Y.~Su, Z.~Meng, S.~Baker, and N.~Collier, ``Few-shot table-to-text generation
  with prototype memory,'' in \emph{Findings of the Association for
  Computational Linguistics: EMNLP 2021}, 2021, pp. 910--917.

\bibitem{reiter1997building}
E.~Reiter and R.~Dale, ``Building applied natural language generation
  systems,'' \emph{Natural Language Engineering}, vol.~3, no.~1, pp. 57--87,
  1997.

\bibitem{liang2009learning}
P.~Liang, M.~I. Jordan, and D.~Klein, ``Learning semantic correspondences with
  less supervision,'' in \emph{Proceedings of the Joint Conference of the 47th
  Annual Meeting of the ACL and the 4th International Joint Conference on
  Natural Language Processing of the AFNLP}, 2009, pp. 91--99.

\bibitem{walker2001spot}
M.~Walker, O.~Rambow, and M.~Rogati, ``Spot: A trainable sentence planner,'' in
  \emph{Second Meeting of the North American Chapter of the Association for
  Computational Linguistics}, 2001.

\bibitem{lu-etal-2009-natural}
W.~Lu, H.~T. Ng, and W.~S. Lee, ``Natural language generation with tree
  conditional random fields,'' in \emph{Proceedings of the 2009 Conference on
  Empirical Methods in Natural Language Processing}.\hskip 1em plus 0.5em minus
  0.4em\relax Singapore: Association for Computational Linguistics, Aug. 2009,
  pp. 400--409.

\bibitem{liu2018table}
T.~Liu, K.~Wang, L.~Sha, B.~Chang, and Z.~Sui, ``Table-to-text generation by
  structure-aware seq2seq learning,'' in \emph{Thirty-Second AAAI Conference on
  Artificial Intelligence}, 2018.

\bibitem{gardent2017webnlg}
C.~Gardent, A.~Shimorina, S.~Narayan, and L.~Perez-Beltrachini, ``The webnlg
  challenge: Generating text from rdf data,'' in \emph{Proceedings of the 10th
  International Conference on Natural Language Generation}, 2017, pp. 124--133.

\bibitem{novikova2017e2e}
J.~Novikova, O.~Dusek, and V.~Rieser, ``The e2e dataset: New challenges for
  end-to-end generation,'' in \emph{18th Annual Meeting of the Special Interest
  Group on Discourse and Dialogue}.\hskip 1em plus 0.5em minus 0.4em\relax
  Association for Computational Linguistics, 2017, pp. 201--206.

\bibitem{see2017get}
A.~See, P.~J. Liu, and C.~D. Manning, ``Get to the point: Summarization with
  pointer-generator networks,'' in \emph{Proceedings of the 55th Annual Meeting
  of the Association for Computational Linguistics (Volume 1: Long Papers)},
  2017, pp. 1073--1083.

\bibitem{elsahar2018zero}
H.~Elsahar, C.~Gravier, and F.~Laforest, ``Zero-shot question generation from
  knowledge graphs for unseen predicates and entity types,'' in
  \emph{Proceedings of NAACL-HLT}, 2018, pp. 218--228.

\bibitem{tseng2018variational}
B.-H. Tseng, F.~Kreyssig, P.~Budzianowski, I.~Casanueva, Y.-C. Wu, S.~Ultes,
  and M.~Gasic, ``Variational cross-domain natural language generation for
  spoken dialogue systems,'' in \emph{Proceedings of the 19th Annual SIGdial
  Meeting on Discourse and Dialogue}, 2018, pp. 338--343.

\bibitem{tran2018dual}
V.-K. Tran and M.~Le~Nguyen, ``Dual latent variable model for low-resource
  natural language generation in dialogue systems,'' in \emph{Proceedings of
  the 22nd Conference on Computational Natural Language Learning}, 2018, pp.
  21--30.

\bibitem{finn2017model}
C.~Finn, P.~Abbeel, and S.~Levine, ``Model-agnostic meta-learning for fast
  adaptation of deep networks,'' in \emph{International conference on machine
  learning}.\hskip 1em plus 0.5em minus 0.4em\relax PMLR, 2017, pp. 1126--1135.

\bibitem{mi2019meta}
F.~Mi, M.~Huang, J.~Zhang, and B.~Faltings, ``Meta-learning for low-resource
  natural language generation in task-oriented dialogue systems,'' in
  \emph{Proceedings of the 28th International Joint Conference on Artificial
  Intelligence}, 2019, pp. 3151--3157.

\bibitem{yao2021knowledge}
H.~Yao, Y.-x. Wu, M.~Al-Shedivat, and E.~Xing, ``Knowledge-aware meta-learning
  for low-resource text classification,'' in \emph{Proceedings of the 2021
  Conference on Empirical Methods in Natural Language Processing}, 2021, pp.
  1814--1821.

\bibitem{su2021plan}
Y.~Su, D.~Vandyke, S.~Wang, Y.~Fang, and N.~Collier, ``Plan-then-generate:
  Controlled data-to-text generation via planning,'' in \emph{Findings of the
  Association for Computational Linguistics: EMNLP 2021}, 2021, pp. 895--909.

\bibitem{houlsby2019parameter}
N.~Houlsby, A.~Giurgiu, S.~Jastrzebski, B.~Morrone, Q.~De~Laroussilhe,
  A.~Gesmundo, M.~Attariyan, and S.~Gelly, ``Parameter-efficient transfer
  learning for nlp,'' in \emph{International Conference on Machine
  Learning}.\hskip 1em plus 0.5em minus 0.4em\relax PMLR, 2019, pp. 2790--2799.

\bibitem{hulora}
E.~J. Hu, P.~Wallis, Z.~Allen-Zhu, Y.~Li, S.~Wang, L.~Wang, W.~Chen
  \emph{et~al.}, ``Lora: Low-rank adaptation of large language models,'' in
  \emph{International Conference on Learning Representations}.

\bibitem{liufew}
H.~Liu, D.~Tam, M.~Mohammed, J.~Mohta, T.~Huang, M.~Bansal, and C.~Raffel,
  ``Few-shot parameter-efficient fine-tuning is better and cheaper than
  in-context learning,'' in \emph{Advances in Neural Information Processing
  Systems}.

\bibitem{hu2021lora}
E.~Hu, Y.~Shen, P.~Wallis, Z.~Allen-Zhu, Y.~Li, S.~Wang, and W.~Chen, ``Lora:
  Low-rank adaptation of large language models,'' 2021.

\bibitem{qin2023modularized}
L.~Qin, X.~Xu, L.~Wang, Y.~Zhang, and W.~Che, ``Modularized pre-training for
  end-to-end task-oriented dialogue,'' \emph{IEEE/ACM Transactions on Audio,
  Speech, and Language Processing}, 2023.

\bibitem{emelin2022injecting}
D.~Emelin, D.~Bonadiman, S.~Alqahtani, Y.~Zhang, and S.~Mansour, ``Injecting
  domain knowledge in language models for task-oriented dialogue systems,''
  \emph{arXiv preprint arXiv:2212.08120}, 2022.

\bibitem{devlin2018bert}
J.~Devlin, M.-W. Chang, K.~Lee, and K.~Toutanova, ``Bert: Pre-training of deep
  bidirectional transformers for language understanding,'' \emph{arXiv preprint
  arXiv:1810.04805}, 2018.

\bibitem{lewis2020bart}
M.~Lewis, Y.~Liu, N.~Goyal, M.~Ghazvininejad, A.~Mohamed, O.~Levy, V.~Stoyanov,
  and L.~Zettlemoyer, ``Bart: Denoising sequence-to-sequence pre-training for
  natural language generation, translation, and comprehension,'' in
  \emph{Proceedings of the 58th Annual Meeting of the Association for
  Computational Linguistics}, 2020, pp. 7871--7880.

\bibitem{ye2023comprehensive}
J.~Ye, X.~Chen, N.~Xu, C.~Zu, Z.~Shao, S.~Liu, Y.~Cui, Z.~Zhou, C.~Gong,
  Y.~Shen \emph{et~al.}, ``A comprehensive capability analysis of gpt-3 and
  gpt-3.5 series models,'' \emph{arXiv preprint arXiv:2303.10420}, 2023.

\bibitem{yang2017anserini}
P.~Yang, H.~Fang, and J.~Lin, ``Anserini: Enabling the use of lucene for
  information retrieval research,'' in \emph{Proceedings of the 40th
  international ACM SIGIR conference on research and development in information
  retrieval}, 2017, pp. 1253--1256.

\bibitem{bialecki2012apache}
A.~Bia{\l}ecki, R.~Muir, G.~Ingersoll, and L.~Imagination, ``Apache lucene 4,''
  in \emph{SIGIR 2012 workshop on open source information retrieval}, 2012,
  p.~17.

\bibitem{wolf2020transformers}
T.~Wolf, L.~Debut, V.~Sanh, J.~Chaumond, C.~Delangue, A.~Moi, P.~Cistac,
  T.~Rault, R.~Louf, M.~Funtowicz \emph{et~al.}, ``Transformers:
  State-of-the-art natural language processing,'' in \emph{Proceedings of the
  2020 conference on empirical methods in natural language processing: system
  demonstrations}, 2020, pp. 38--45.

\bibitem{DBLP:journals/corr/KingmaB14}
\BIBentryALTinterwordspacing
D.~P. Kingma and J.~Ba, ``Adam: {A} method for stochastic optimization,'' in
  \emph{3rd International Conference on Learning Representations, {ICLR} 2015,
  San Diego, CA, USA, May 7-9, 2015, Conference Track Proceedings}, Y.~Bengio
  and Y.~LeCun, Eds., 2015. [Online]. Available:
  \url{http://arxiv.org/abs/1412.6980}
\BIBentrySTDinterwordspacing

\bibitem{ma2019key}
S.~Ma, P.~Yang, T.~Liu, P.~Li, J.~Zhou, and X.~Sun, ``Key fact as pivot: A
  two-stage model for low resource table-to-text generation,'' in
  \emph{Proceedings of the 57th Annual Meeting of the Association for
  Computational Linguistics}, 2019, pp. 2047--2057.

\bibitem{raffel2020exploring}
C.~Raffel, N.~Shazeer, A.~Roberts, K.~Lee, S.~Narang, M.~Matena, Y.~Zhou,
  W.~Li, P.~J. Liu \emph{et~al.}, ``Exploring the limits of transfer learning
  with a unified text-to-text transformer.'' \emph{J. Mach. Learn. Res.},
  vol.~21, no. 140, pp. 1--67, 2020.

\bibitem{zhao-etal-2021-attend-memorize}
\BIBentryALTinterwordspacing
W.~Zhao, Y.~Liu, Y.~Wan, and P.~Yu, ``Attend, memorize and generate: Towards
  faithful table-to-text generation in few shots,'' in \emph{Findings of the
  Association for Computational Linguistics: EMNLP 2021}.\hskip 1em plus 0.5em
  minus 0.4em\relax Punta Cana, Dominican Republic: Association for
  Computational Linguistics, Nov. 2021, pp. 4106--4117. [Online]. Available:
  \url{https://aclanthology.org/2021.findings-emnlp.347}
\BIBentrySTDinterwordspacing

\bibitem{wu2019response}
Y.~Wu, F.~Wei, S.~Huang, Y.~Wang, Z.~Li, and M.~Zhou, ``Response generation by
  context-aware prototype editing,'' in \emph{Proceedings of the AAAI
  Conference on Artificial Intelligence}, vol.~33, no.~01, 2019, pp.
  7281--7288.

\bibitem{papineni2002bleu}
K.~Papineni, S.~Roukos, T.~Ward, and W.-J. Zhu, ``Bleu: a method for automatic
  evaluation of machine translation,'' in \emph{Proceedings of the 40th annual
  meeting of the Association for Computational Linguistics}, 2002, pp.
  311--318.

\bibitem{lin2004rouge}
C.-Y. Lin, ``Rouge: A package for automatic evaluation of summaries,'' in
  \emph{Text summarization branches out}, 2004, pp. 74--81.

\bibitem{dhingra2019handling}
B.~Dhingra, M.~Faruqui, A.~Parikh, M.-W. Chang, D.~Das, and W.~Cohen,
  ``Handling divergent reference texts when evaluating table-to-text
  generation,'' in \emph{Proceedings of the 57th Annual Meeting of the
  Association for Computational Linguistics}, 2019, pp. 4884--4895.

\bibitem{radford2019language}
A.~Radford, J.~Wu, R.~Child, D.~Luan, D.~Amodei, I.~Sutskever \emph{et~al.},
  ``Language models are unsupervised multitask learners.''

\bibitem{pfeiffer2020AdapterHub}
\BIBentryALTinterwordspacing
J.~Pfeiffer, A.~R\"uckl\'{e}, C.~Poth, A.~Kamath, I.~Vuli\'{c}, S.~Ruder,
  K.~Cho, and I.~Gurevych, ``Adapterhub: A framework for adapting
  transformers,'' in \emph{Proceedings of the 2020 Conference on Empirical
  Methods in Natural Language Processing (EMNLP 2020): Systems
  Demonstrations}.\hskip 1em plus 0.5em minus 0.4em\relax Online: Association
  for Computational Linguistics, 2020, pp. 46--54. [Online]. Available:
  \url{https://www.aclweb.org/anthology/2020.emnlp-demos.7}
\BIBentrySTDinterwordspacing

\end{thebibliography}

{\appendices
\begin{figure*}[!t]
\centering
\includegraphics[scale=0.54]{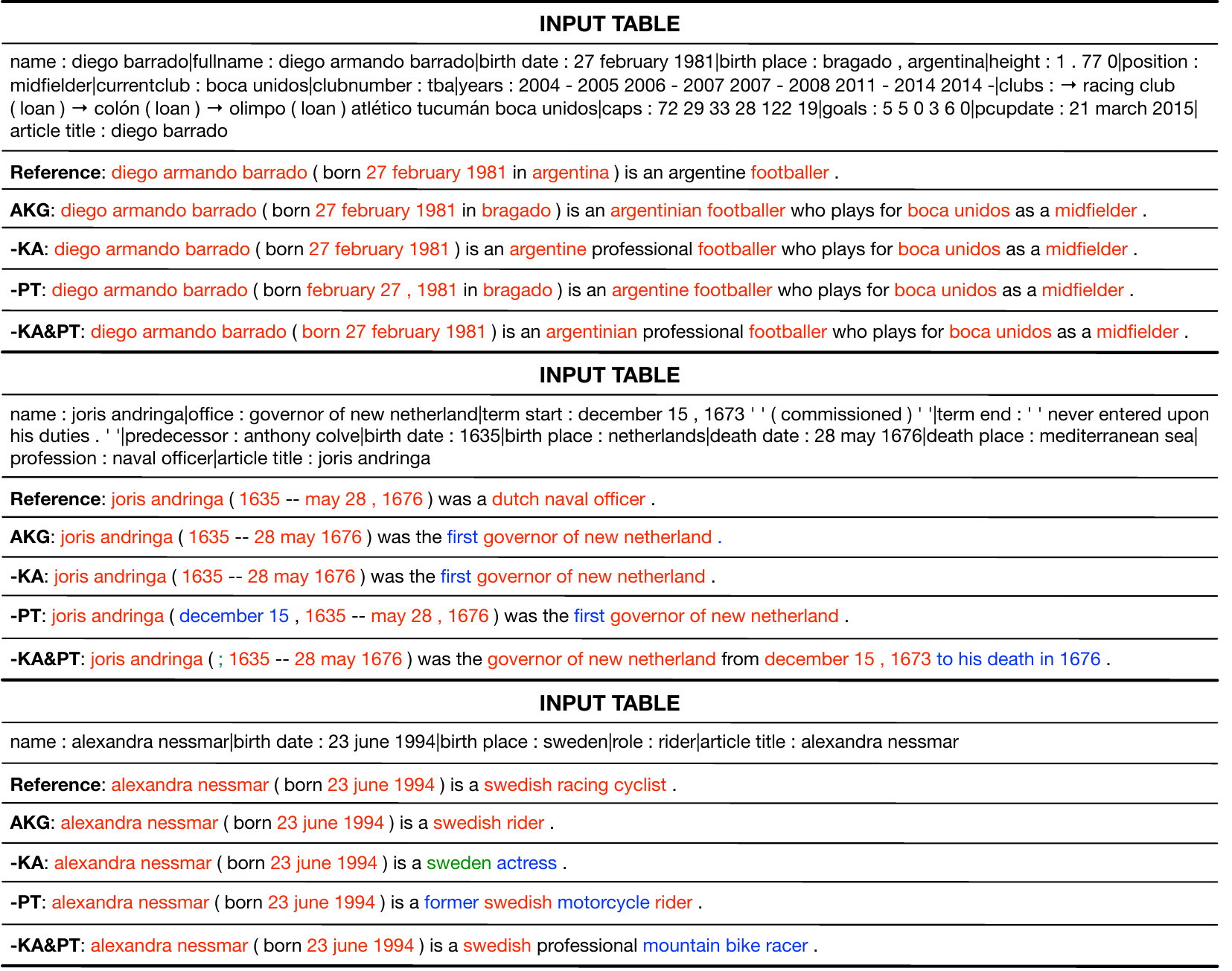}
\caption{Additional examples of generated results from the ablation study. Red denotes information supported by the tabular data, blue denotes information that contradicts the tabular data, and green denotes grammatical mistakes that influence fluency.}
\label{Figure6}
\end{figure*}

\section{Experiment Details}
We apply Adam \cite{DBLP:journals/corr/KingmaB14} as our optimizer with the learning rate set to 0.00003. The mini-batch size is set to 16. We use BART-large \cite{lewis2020bart} as our backbone generator with the Hugging Face Library \cite{wolf2020transformers} with default settings. We set $<sep>$, $<eos>$, and $<context\_start>$ as special tokens to the BART vocabulary throughout the templating process. Moreover, we insert the knowledge adapter, based on the AdapterHub Library \cite {pfeiffer2020AdapterHub}, into each encoder and decoder layer. For more details, refer to our released code and data at \url{https://github.com/sjtugzx/PromptMize.git}.

\section{Details of the Human Evaluation Setup}
To perform the human evaluation, we randomly select 100 generated sentences of the closest baseline, P2G (BART-large), and AKG with the corresponding table--reference pairs from the Humans data set with 100 paring instances. To reduce human bias, we randomly shufﬂe these 200 samples before presenting them to the three annotators. All volunteers were postgraduate students with NLP-related knowledge. Following previous research \cite{chen2020few}, we evaluate each generated sentence with two tasks: evaluating faithfulness and language fluency. Throughout the evaluation, all annotators are asked to follow the annotation guidelines. \newline
\textbf{Human Evaluation Guidelines} \newline
Give original tabular data and the generated descriptions. Annotators are asked to annotate and statistic the following three tasks:
\begin{itemize}
\item \#Sup: Count the content supported by the tabular data.
\item \#Cont: Count the content contradicting the tabular data.
\item Fluency: Estimate the fluency of the generated sentences. (Ignore the faithfulness, the sentence is fluent if it is grammatical and natural.) Likert scale is 1, 2, 3. 1 denotes the sentences containing obvious grammatical errors or are poorly formed. 2 denotes the sentences flow smoothly, with problems that do not affect the reading. 3 denotes that the sentences are fluent without any mistakes.
\end{itemize} 
\textbf{\textit{Table Example}}: \newline
Tabular Data: \newline
name: michael phillip wojewoda \newline
image: michael phillip wojewoda . jpg \newline
caption: wojewoda performing with the rheostatics in 2007 at massey hall \newline
background: nonvocalinstrumentalist \newline
origin: toronto , ontario , canada \newline
genre: indie rock \newline
occupation: musician , record producer \newline
associated acts: rheostatics space invaders \newline
article title: michael phillip wojewoda \newline
Generated Sentences: michael phillip wojewoda is an indie rock musician , record producer , and guitarist . \newline
\textbf{\textit{Ranking Example}}:
\begin{itemize}
\item \#Sup: 4. Reason: michael phillip wojewoda, indie rock, musician, record producer are supported by the tabular data.
\item \#Cont: 1. Reason: guitarist is not supported by the tabular data.
\item Fluency: 2. Reason: The overall expression is fluent. There are no grammatical errors, and any mistakes affect the human reading.
\end{itemize}

\section{More Examples of Generated Results}
In this part, we provide more generated examples from the ablation study. The generated results are shown in Fig.~\ref{Figure6}. From the results, we can see that our model can generate ﬂuent and diverse sentences. The prompt planner and the knowledge adapter improve the generation fluency and faithfulness to different extents compared with the baseline model BART-large. After applying all techniques, we can see that the generated quality is further improved. These results further demonstrate the applicability and generalization ability of our model.

}


\end{document}